\documentclass{article}

    \PassOptionsToPackage{numbers, compress}{natbib}

\usepackage[preprint]{unireps_2023}

\usepackage{amsmath,amsfonts,bm}

\def\eqref#1{equation~\ref{#1}}

\def\1{\bm{1}}

\def\vx{{\bm{x}}}
\def\vy{{\bm{y}}}

\DeclareMathAlphabet{\mathsfit}{\encodingdefault}{\sfdefault}{m}{sl}
\SetMathAlphabet{\mathsfit}{bold}{\encodingdefault}{\sfdefault}{bx}{n}

\DeclareMathOperator*{\argmax}{arg\,max}

\usepackage[colorlinks,linkcolor={blue!50!black},citecolor={blue!50!black},urlcolor={blue!50!black},backref=page,bookmarks=false]{hyperref}       %
\usepackage{url}
\usepackage{tabularx}

\usepackage[labelformat=simple]{subcaption}
 
\usepackage{tikz}
\usetikzlibrary{bayesnet}

\usepackage{bm}
\usepackage{xspace}
\usepackage{amsmath}
\usepackage{amssymb}
\usepackage{mathtools}
\usepackage{amsthm}

\usepackage{booktabs}
\usepackage{multirow}
\usepackage{wasysym}
\usepackage{wrapfig}

\usepackage[textsize=tiny]{todonotes}

\usepackage[nameinlink]{cleveref}
\crefname{equation}{eq.}{eqs.}
\Crefname{equation}{Eq.}{Eqs.}
\usepackage{mathtools}

\usepackage{tcolorbox}

\graphicspath{{./figs/}}

\newcolumntype{Y}{>{\centering\arraybackslash}X}

\newcommand{\beq}{\begin{equation}}
\newcommand{\eeq}{\end{equation}}
\newcommand{\Ell}{{\mathcal L}}

\newcommand{\g}{\,|\,}

\definecolor{purp}{HTML}{7E1E9C}

\definecolor{baseline}{HTML}{1f77b4}
\definecolor{dmaapx}{HTML}{2ca02c}
\definecolor{aug}{HTML}{9467bd}

\newtcolorbox{remark}[1]{
  colback=gray!5!white,
  colframe=gray!75!black,
  title={\textbf{Remark} (#1)},
  rounded corners,
  boxrule=1pt,
  fonttitle=\normalfont
}

\usepackage{colortbl}
\definecolor{Gray2}{gray}{0.94}
\definecolor{Gray}{gray}{0.8}
\newcolumntype{a}{>{\columncolor{Gray}}c}

\newlength\savewidth\newcommand\shline{\noalign{\global\savewidth\arrayrulewidth
  \global\arrayrulewidth 1pt}\hline\noalign{\global\arrayrulewidth\savewidth}}

\usepackage{pifont}%

\title{A Compact Representation for Bayesian Neural Networks By Removing Permutation Symmetry}

\author{
Tim Z. Xiao\textsuperscript{1,2} \\
\texttt{\small zhenzhong.xiao@uni-tuebingen.de} \\
\And
Weiyang Liu\textsuperscript{3,4} \\
\texttt{\small wl396@cam.ac.uk} \\
\And
Robert Bamler\textsuperscript{1} \\
\texttt{\small robert.bamler@uni-tuebingen.de} \\
\AND
\small\textsuperscript{1}University of T\"ubingen ~~~~\small\textsuperscript{2}IMPRS-IS ~~~~\small\textsuperscript{3}University of Cambridge \\
\small\textsuperscript{4}Max Planck Institute for Intelligent Systems, Tübingen
}

\begin{document}

\maketitle

\begin{abstract}
Bayesian neural networks (BNNs) are a principled approach to modeling predictive uncertainties in deep learning, which are important in safety-critical applications.
Since exact Bayesian inference over the weights in a BNN is intractable, various approximate inference methods exist, among which sampling methods such as Hamiltonian Monte Carlo (HMC) are often considered the gold standard.
While HMC provides high-quality samples, it lacks interpretable summary statistics because its sample mean and variance is meaningless in neural networks due to permutation symmetry.
In this paper, we first show that the role of permutations can be meaningfully quantified by a number of transpositions metric.
We then show that the recently proposed rebasin method~\citep{ainsworth2022git} allows us to summarize HMC samples into a compact representation that provides a meaningful explicit uncertainty estimate for each weight in a neural network, thus unifying sampling methods with variational inference.
We show that this compact representation allows us to compare trained BNNs directly in weight space across sampling methods and variational inference, and to efficiently prune neural networks trained without explicit Bayesian frameworks by exploiting uncertainty estimates from HMC.
\end{abstract}

\section{Introduction and Background}
\label{sec:intro}

\begin{wrapfigure}{r}{0.3\textwidth}
  \vspace{-2.5ex}
  \centering
  \includegraphics[width=0.29\textwidth]{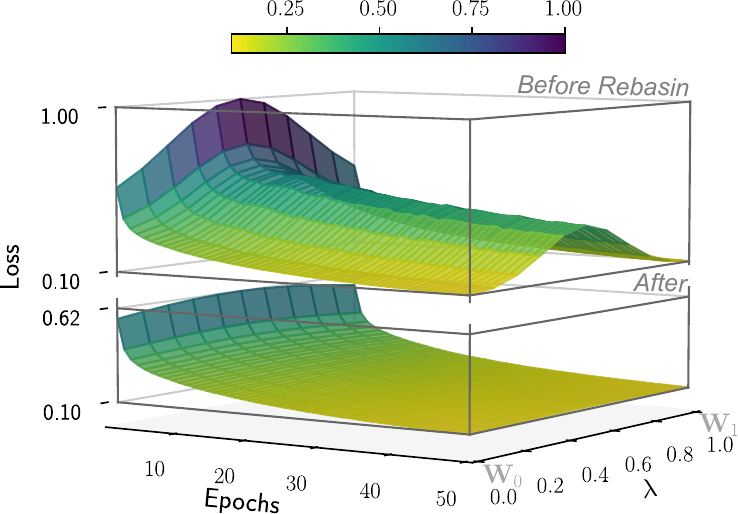}
  \caption{Training dynamics for models with $\mathbf{W}_0$ and $\mathbf{W}_1$, and their interpolations~$\mathbf W_\lambda$.}
  \vspace{-1ex}
  \label{fig:training_dynamics_loss}
\end{wrapfigure}

When training a neural network on a data set~$\mathcal D$, one minimizes a loss function $\mathcal L(\mathbf W, \mathcal D)$ over weights and biases (collectively referred to as ``weights'' and denoted as boldface~$\mathbf W$ in the following).
Yet, when comparing two trained networks (e.g., to choose a training algorithm or learning rate), one rarely compares the trained weights directly.
Instead, one compares various performance metrics of the two trained networks on a held-out data set.
Direct comparisons in weight space are difficult, in part because standard distance metrics (e.g., the euclidean distance) are not meaningful in weight space due to a permutation symmetry of neural networks \citep{hecht1990algebraic, brea2019weight, entezari2021role, ainsworth2022git}:
consider two consecutive layers of a neural network, which represent a function $x \mapsto \sigma_2(W_2  \sigma_1(W_1 x + b_1) + b_2)$, where $W_{1,2}$, $b_{1,2}$, and $ \sigma_{1,2}$ are weight matrices, bias vectors, and (componentwise applied) activation functions, respectively.
It is easy to see that this function does not change if we consider an arbitrary permutation matrix~$P$ and replace the weights and biases by $W_1':= PW_1$, $b_1' := Pb_1$, and $W_2' := W_2P^{-1}$.
Thus, euclidean distances in weight space like $||W_1 - W_1'||_2^2$ are not meaningful.

Recent works~\citep{ainsworth2022git,entezari2021role} propose and analyze a method called rebasin, which efficiently finds a permutation that brings the weights~$\mathbf W_{1}$ of one neural network as close as possible to the weights~$\mathbf W_{0}$ of an independently trained reference network with the same architecture.
The authors show that applying this permutation essentially removes the loss barrier~ \citep{frankle2020linear} when linearly interpolating between the weights of the two neural networks (see \Cref{fig:training_dynamics_loss} for before and after rebasin, where $\lambda$~controls the interpolation $\mathbf W_{\lambda} := (\lambda-1)\mathbf W_{0} + \lambda \mathbf W_{1}$).
The authors conjecture that the loss landscape is quasi-convex up to permutations.
In this paper, we build on these findings in three ways
\begin{enumerate}
  \item
    We propose to quantify rebasin permutations by their number of transpositions (NoT), see below.
    We show empirically that NoT is a valuable metric for analyzing weight-space geometry as it is remarkably stable over training and correlates strongly with the loss barrier.
  \item
    We argue that a quasi-convex loss landscape is particularly useful for Bayesian neural networks (BNNs), which aim to find (an approximation of) the posterior distribution $p({\mathbf W \g \mathcal D})$ over weights~$\mathbf W$ that are consistent with the data~$\mathcal D$.
    We show empirically that rebasin allows us to summarize the approximate posterior of sampling based inference methods like Hamiltonian Monte Carlo (HMC) in the same compact representation that variational inference (VI) uses, thus enabling direct comparisons across inference methods.
  \item
    We show that the proposed unifying compact representation of BNNs is interpretable.
    For example, unusual for sampling methods in BNNs, we obtain meaningful explicit uncertainty estimates in weight space from HMC.
    We show that we can use these uncertainty estimates from HMC to efficiently prune a neural network trained with deep ensembles~\citep{lakshminarayanan2017simple}.
\end{enumerate}

The paper is structured as follows:
we introduce and empirically analyzes the NoT metric (item~1 above) in \Cref{sec:not}, discuss BNNs (items 2 and~3 above) in \Cref{sec:bnn}, and conclude in \Cref{sec:conclusions}.

\section{Quantifying Permutations in Weight Space by Number of Transpositions}
\label{sec:not}

\begin{figure}[t]
  \centering
  \includegraphics[width=\textwidth]{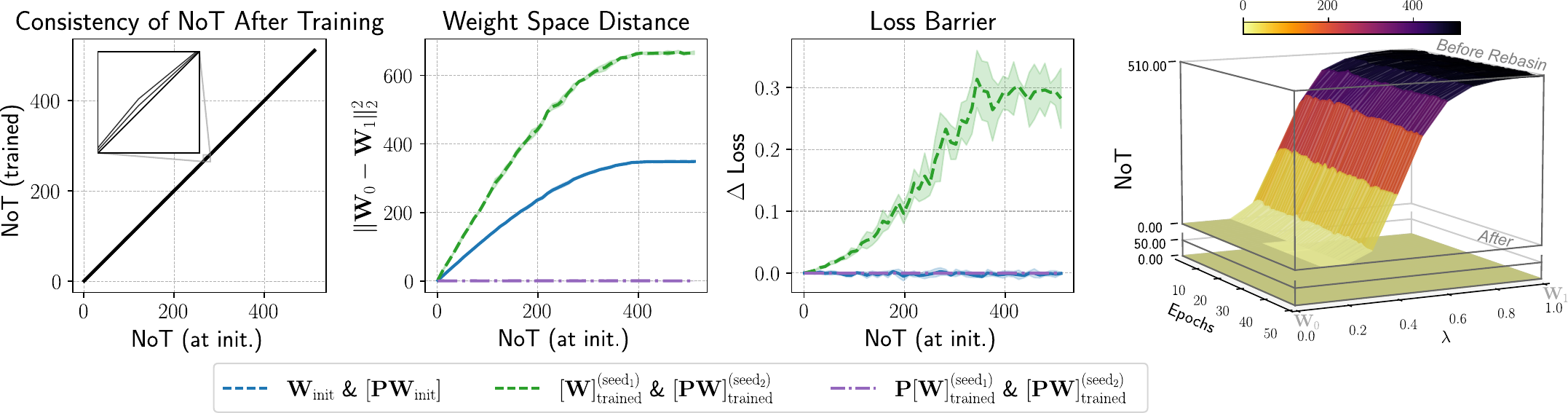}
  \caption{
    Left three: effect of permuting initial weights by different Number of Transpositions (NoT) on NoT after training, weight-space distance, and loss barrier (shaded regions: $\pm1\sigma$~over 5~runs).
    Right: NoT changes monotonically along the interpolation $\mathbf W_\lambda$ between two models $\mathbf{W}_0$ and $\mathbf{W}_1$.
  }
  \label{fig:nots_vs_wdiff_bars}
\end{figure}

We briefly introduce and analyze a metric to quantify the ``magnitude'' of permutations obtained from rebasin~\citep{ainsworth2022git}, which proved useful for building up intuition and for debugging implementations of experiments discussed in \Cref{sec:bnn}.
Readers only interested in BNNs may opt to skip this part.

We propose to measure the magnitude of a permutation by its number of transpositions (NoT), i.e., the minimal number of pairwise swaps whose consecutive execution results in the given permutation.
It is a well-known result from algebra~\citep{knapp2007basic} that this is always possible, and that NoT can be calculated efficiently by first factorizing a permutation into non-overlapping cycles and then expressing each cycle of length~$k$ as a product of $(k-1)$ transpositions.
For example, the permutation~$P$ that maps $1\mapsto4$, $2\mapsto1$, $3\mapsto5$, $4\mapsto2$, and $5\mapsto3$ can be written as $P = {({1\;4\;2})\,({3\;5})} = {({1\;4})\, ({4\;2})\,({3\;5})}$, where $({a_1,a_2,\ldots,a_k})$ denotes a cycle $a_1\mapsto a_2 \mapsto \cdots \mapsto a_k \mapsto a_1$.
Thus, $\operatorname{NoT}(P)=3$.

We find empirically that NoT is a meaningful metric for analyzing weight space geometry.
We consider a neural network for classification of MNIST digits with a single hidden layer and a total of 512 hidden activations that can be permuted.
We trained a network with randomly initialized weights $\mathbf W_\mathrm{init}$ and a set of networks whose weights were initialized as $P\mathbf W_\mathrm{init}$ where $P$~is a random permutation with $\operatorname{NoT}(P)$ ranging over all values from zero to~511.
After training both neural networks, we obtained a permutation~$P'$ by matching the two trained networks using rebasin.

As a first high-level check, \Cref{fig:nots_vs_wdiff_bars}~(left) shows $\operatorname{NoT}(P')$ as a function of $\operatorname{NoT}(P)$ for three random~$\mathbf W_\mathrm{init}$.
Even though the permuted and unpermuted networks were trained with different random seeds for the sampling of minibatches, NoT remains almost exactly unaffected by training.
Further, \Cref{fig:nots_vs_wdiff_bars}~(center two) show that both the euclidean distance in weight space and the barrier of the loss function after training correlate strongly with NoT.
Here, the barrier is defined as in \citep{frankle2020linear} as $\max_{\lambda \in [0,1]} \Ell(\mathbf W_\lambda, \mathcal D) - \frac{1}{2}\big(\Ell(\mathbf{W}_0,\mathcal D) + \Ell(\mathbf{W}_1,\mathcal D)\big)$, where $\mathbf W_{\lambda} := (\lambda-1)\mathbf W_{0} + \lambda \mathbf W_{1}$, and $\mathbf W_0$~and $\mathbf W_1$ are the weights of the two trained models.
Finally, \Cref{fig:nots_vs_wdiff_bars}~(right) analyzes $\operatorname{NoT}(\tilde P)$ both over training epochs and along the linear interpolation~$\lambda\in[0,1]$, where $\tilde P$ is obtained by matching $\mathbf W_\lambda$ to $\mathbf W_0$ using rebasin.
We observe that the $\operatorname{NoT}(\tilde P)$ remains flat in the vicinity of either trained model, and changes smoothly and monotonically in between.

From these observations, we conclude that, while comparing neural network weights by euclidean distance alone is not meaningful (see \Cref{sec:intro}), we can meaningfully quantify weight-space distances by a pair $\big(||\mathbf W_{\!0} - P\mathbf W_{\!1}||_2^2, \operatorname{NoT}(P)\big)$ where the permutation~$P$ matches $\mathbf W_{\!1}$ to~$\mathbf W_{\!0}$ by rebasin.

\section{A Unifying Compact Representation for Bayesian Neural Networks}
\label{sec:bnn}

Building on the argument \citep{pourzanjani2017improving, kurle2021symmetries, wiese2023towards} that removing permutation degrees of freedom is particularly useful for \emph{Bayesian} Neural Networks (BNNs), we propose a framework to combine the strengths of two classes of inference algorithms in BNNs.
Training a BNN amounts to finding (an approximation of) the so-called posterior distribution $p({\mathbf W \g \mathcal D})$ of all weights~$\mathbf W$ that are consistent with the training data~$\mathcal D$, and thus involves more than a single set of weights.
We show below that being able to meaningfully compare weights makes approximate posteriors of BNNs more interpretable.

The exact posterior distribution is $p({\mathbf W \g \mathcal D}) = p(\mathbf W) p({\mathcal D \g \mathbf W}) / p(\mathcal D)$, where $p(\mathbf W)$ is a prior distribution that acts like a regularizer (often an isotropic Gaussian), $p({\mathcal D \g \mathbf W}) = \exp[-\mathcal L(\mathbf W,\mathcal D)]$, and $p(\mathcal D) = \int p(\mathbf W) p({\mathcal D \g \mathbf W}) \,\mathrm{d}\mathbf W$.
The exact posterior is usually intractable in BNNs, but various efficient approximation methods have been developed.
We group them into two categories:
\begin{enumerate}
  \item[(a)] \textbf{Parametric methods},
    such as variational inference (VI; \citep{blundell2015weight}) and Laplace approximation~\citep{ritter2018scalable, kristiadi2020being} approximate the posterior $p(\mathbf{W} \g \mathcal{D})$ explicitly with a simpler distribution
    $q(\mathbf{W})$, e.g., a fully factorized normal distribution with fitted means and variances;
  \item[(b)] \textbf{Sampling methods},
    such as Hamiltonian Monte Carlo (HMC;~\citep{neal2011mcmc}), stochastic gradient Langevin dynamics~\citep{welling2011bayesian}, and MCDropout \citep{gal2016dropout} draw a set of $K$~samples $\{\mathbf{W}^{(k)}\}_{k=1}^K$ directly from $p(\mathbf{W} \g \mathcal{D})$ without explicit representing their distribution;
    deep ensembles~\citep{lakshminarayanan2017simple} is sometimes also considered in this context despite not following the Bayesian framework.
\end{enumerate}
As sampling methods~(b) lack explicit uncertainty information, one might be tempted to fit samples with a parametric distribution, e.g., a Gaussian $q_{\mathrm{d}}(\mathbf{W}) = \mathcal{N}(\bm{\mu}_{\mathrm{d}}, \mathrm{diag}(\bm{\sigma}^2_{\mathrm{d}}))$, where $\bm{\mu}_{\mathrm{d}}$ and~$\bm{\sigma}^2_{\mathrm{d}}$ are the sample mean and variance, and the subscript `d' is for `direct'.
However, we show in \Cref{sec:agreement-tv} that $q_\mathrm{d}$~is a poor posterior approximation, likely because the posterior is multimodal due to the permutation symmetry.
However, the quasi-convexity conjecture~\hbox{\citep{ainsworth2022git,entezari2021role}} suggests that the posterior is unimodal once we remove the permutation degrees of freedom.
Therefore, we propose to summarize samples from, e.g., HMC, with a diagonal Gaussian $q_{\mathrm{r}}(\mathbf{W}) = \mathcal{N}(\bm{\mu}_{\mathrm{r}}, \mathrm{diag}(\bm{\sigma}^2_{\mathrm{r}}))$ where $\bm{\mu}_{\mathrm{r}}$ and~$\bm{\sigma}^2_{\mathrm{r}}$ are the sample mean and variance after using rebasin~\citep{ainsworth2022git} to match each sample to an arbitrary shared reference sample.
We show in \Cref{sec:agreement-tv} that $q_\mathrm{r}$~approximates the posterior well despite its compactness, and in \Cref{sec:comparisons-weighspace} that having a unifying compact representation allows us to combine the respective strengths of different inference methods, thus going beyond the findings in \citep{wiese2023towards}.

All experiments were done with a simple fully connected network for MNIST classification with a single hidden layer of size~512.
We compare $q_\mathrm{d}$ and~$q_\mathrm{r}$ across HMC, deep ensemble, and VI.

\subsection{Compact Representation as an Approximate Posterior}
\label{sec:agreement-tv}

\begin{table}[t]
\caption{
\footnotesize Performance of different BNNs ($q_{\mathrm{d}}$: before rebasin; $q_{\mathrm{r}}$: after rebasin) on their agreement~(\Cref{eq:agreement}) and total variation (TV; \Cref{eq:tv}) to HMC samples, and on their test set accuracy.
}
 \footnotesize
 \centering
 \setlength{\tabcolsep}{3pt}
 \renewcommand{\arraystretch}{1.2}
 \scalebox{0.9}{
 \label{tab:agree_tv}
 \begin{tabular}{l|ccc|ccc|c}
    & \multicolumn{3}{c|}{HMC} & \multicolumn{3}{c|}{Ensemble} & VI \\
    & Sample & $q_{\mathrm{d}}(\mathbf{W})$ & $q_{\mathrm{r}}(\mathbf{W})$ & Sample & $q_{\mathrm{d}}(\mathbf{W})$ & $q_{\mathrm{r}}(\mathbf{W})$ & $q(\mathbf{W})$  \\
    \shline
    ($\uparrow$) Agreement with HMC samples     & $1.$ & \cellcolor{Gray2}{$0.1212$} & \cellcolor{Gray}{$0.8249$} & $0.9931$ & \cellcolor{Gray2}{$0.5239$} & \cellcolor{Gray}{$0.9868$}  & $0.9885$ \\
    ($\downarrow$) TV to HMC samples        & $0.$ & \cellcolor{Gray2}{$0.8641$} & \cellcolor{Gray}{$0.6570$} & $0.0229$ & \cellcolor{Gray2}{$0.7210$} & \cellcolor{Gray}{$0.0495$}  & $0.0235$ \\
    \shline
    Test Accuracy (\%) of Samples     & $98.43$ & \cellcolor{Gray2}{$11.11$} & \cellcolor{Gray}{$82.34$} & $98.66$ & \cellcolor{Gray2}{$52.25$} & \cellcolor{Gray}{$97.72$}  & $98.11$ \\
    Test Accuracy (\%) of $\bm{\mu}_{\mathrm{d}}$ and $\bm{\mu}_{\mathrm{r}}$       & N/A & \cellcolor{Gray2}{$28.06$} & \cellcolor{Gray}{$92.25$} & N/A & \cellcolor{Gray2}{$86.40$} & \cellcolor{Gray}{$97.97$}  & $98.04$ \\
 \end{tabular}}
\end{table}

Before we use $q_{\mathrm{r}}(\mathbf{W})$ to compare models directly in weight space, we first evaluate whether it provides a good approximation of the posterior despite radically reducing the amount of information available in the samples.
Evaluations of BNNs are typically done by comparing the predictive distribution $p(\vy^* \g \vx^*, \mathcal{D}) = \int p(\vy^* \g \vx^*, \mathbf{W}) p(\mathbf{W} \g \mathcal{D}) \,\mathrm{d} \mathbf{W}$ on a test set $\mathcal D_\mathrm{test}$, where $\mathbf x^*$~is the input, $\mathbf y^*$~is the prediction, and we assume that $p(\vy^* \g \vx^*, \mathbf W)$ is described by the neural network.
There are two popular metrics to compare predictive distributions between a method~$p$ and HMC (which is often considered the most precise approximation of a BNN posterior \citep{izmailov2021bayesian}):
\emph{agreement} and \emph{total variation} (TV) \citep{wilson2022evaluating} (in the following, $I[\,\cdot\,]$ is the indicator function),
\begin{align}
  \operatorname{Agree.}(p, p_\mathrm{HMC}) & = \frac{1}{|\mathcal D_\mathrm{test}|} \sum_{\vx^*\in\mathcal D_\mathrm{test}} \!\!\!\!I\big[\argmax_{\vy^*} p(\vy^* \g \vx^*, \mathcal{D}) = \argmax_{\vy^*} p_\mathrm{HMC}(\vy^* \g \vx_i^*, \mathcal{D})\big]; \label{eq:agreement} \\
  \operatorname{TV}(p, p_\mathrm{HMC}) & = \frac{1}{|\mathcal D_\mathrm{test}|} \sum_{\vx^*\in\mathcal D_\mathrm{test}} \!\!\!\!\frac{1}{2} \sum_{\vy^*} \big|p(\vy^* \g \vx_i^*, \mathcal{D}) - p_\mathrm{HMC}(\vy^* \g \vx_i^*, \mathcal{D}) \big|. \label{eq:tv}
\end{align}
Results in \Cref{tab:agree_tv} show that: 
(i)~$q_{\mathrm{r}}(\mathbf{W})$ has much better performance than $q_{\mathrm{d}}(\mathbf{W})$ for both HMC and ensemble; thus, rebasin is crucial for obtaining a accurate compact representation;
(ii)~ensemble outperforms HMC in $q_{\mathrm{r}}(\mathbf{W})$, which could indicate issues of the activation matching algorithm~\citep{ainsworth2022git} when applied between networks with different loss levels, e.g., within HMC samples;
and (iii)~$q_{\mathrm{r}}(\mathbf{W})$ provides a parameter efficient representation for ensemble with competitive performance.

\subsection{Comparing and Merging BNNs in Weight Space}
\label{sec:comparisons-weighspace}

\begin{figure}[t]
  \centering
  
  \begin{subfigure}[t]{0.43\textwidth}
    \centering
    \includegraphics[width=\linewidth]{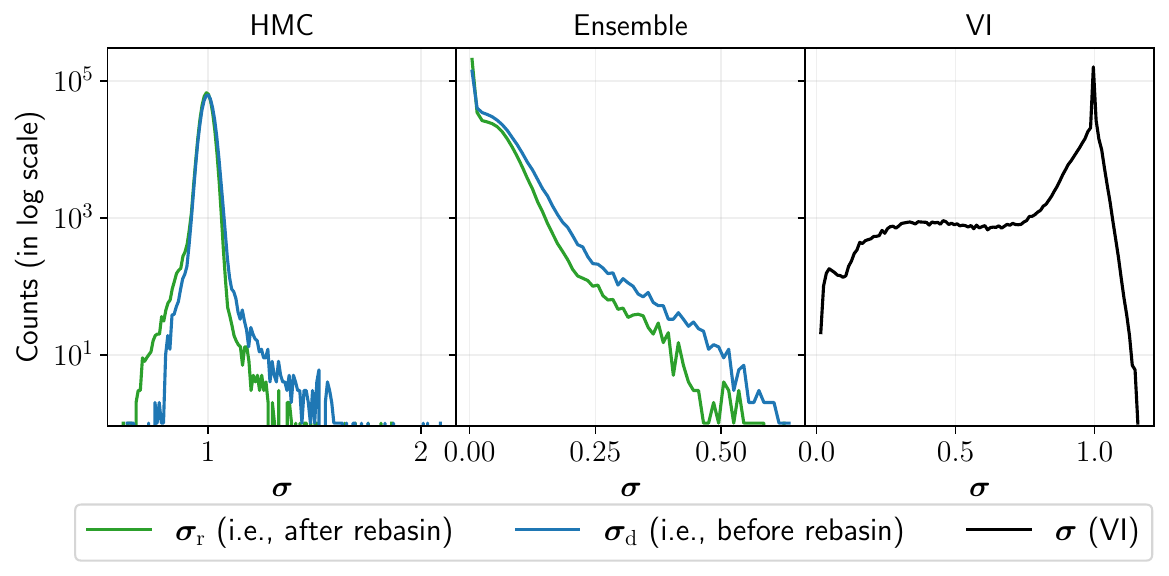}
    \label{fig:sig_hist}
  \end{subfigure}
  \hfill
  \begin{subfigure}[t]{0.52\textwidth}
    \centering
    \includegraphics[width=\linewidth]{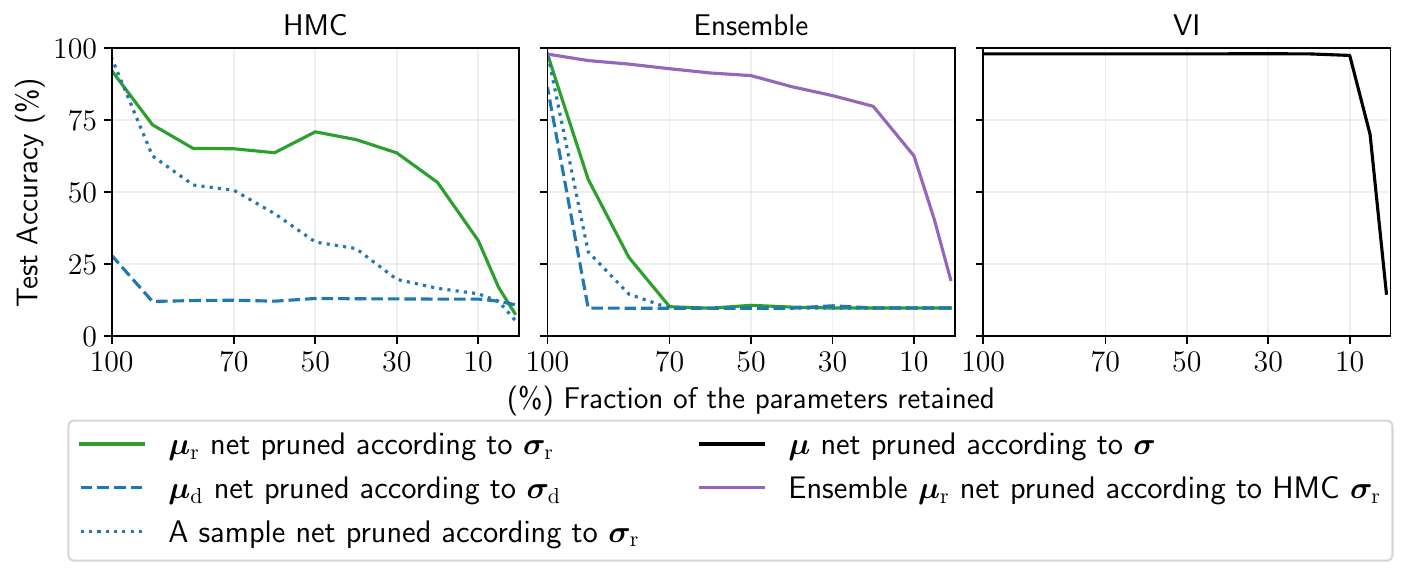}
    \label{fig:pruning}
  \end{subfigure}
  
  \caption{
  Left: histograms of the standard deviation $\bm{\sigma}$~of weights before ($\bm{\sigma}_{\mathrm{d}}$) and after ($\bm{\sigma}_{\mathrm{r}}$) rebasin. 
  Right: test accuracy vs.\ various levels of weight pruning (retaining only weights with lowest~$\bm{\sigma}$).}
  \label{fig:hist_and_pruning}
\end{figure}

\Cref{fig:hist_and_pruning} (left) shows histograms of the variances $\bm{\sigma}_\mathrm{d}$ and~$\bm{\sigma}_\mathrm{r}$ for HMC and ensemble, and the variances fitted by VI.
We observe that (i)~permuting all samples into the same basin reduces variances overall, as expected;
and (ii)~the two Bayesian methods (HMC and VI) have lots of weights with a variance close to one, which is not affected by rebasin in HMC.
This indicates mode collapse since we used a standard Gaussian prior, i.e., the Bayesian methods identify these weights as unnecessary.

The proposed unifying compact representation allows us to merge different BNNs by stitching the means $\bm{\mu}_{\mathrm{r}}$ from one model with the variances $\bm{\sigma}_{\mathrm{r}}^2$ from another.
\Cref{fig:hist_and_pruning} (right) shows test accuracies of neural networks after pruning weights with high~$\bm{\sigma}_\mathrm{r}^2$ (a simplified variant of the compression method in~\citep{yang2020variational,tan2022posttraining}).
The purple curve uses weights~$\bm{\mu}_\mathrm{r}$ from ensemble but $\bm{\sigma}_\mathrm{r}^2$ from HMC, thus combining the predictive strength of ensemble with the accurate uncertainty estimates of HMC.
It significantly outperforms both variants that use only the ensemble or only HMC (green curves).

\section{Conclusion}
\label{sec:conclusions}

When doing Bayesian inference in BNNs, it is straightforward to go from parametric based to sampling based inference, e.g., one can easily draw samples from a variational distribution.
But permutation symmetry makes it difficult to go in the reverse direction.
In this work, we use the recently proposed rebasin method to remove the permutation symmetry.
We propose a unifying compact representation for Bayesian inference in BNNs, which allows us to go from sampling based inference to parametric based inference, and to combine the respective strengths of different inference methods.

\newpage
\subsubsection*{Acknowledgments}
The authors would like to thank Yingzhen Li, Takeru Miyato, Johannes Zenn, Nicolò Zottino and Andi Zhang for helpful discussions.
Funded by the Deutsche Forschungsgemeinschaft (DFG, German Research Foundation) under Germany’s Excellence Strategy~--~EXC number 2064/1~--~Project number 390727645.
This work was supported by the German Federal Ministry of Education and Research (BMBF): Tübingen AI Center, FKZ:~01IS18039A.
Robert Bamler acknowledges funding by the German Research Foundation (DFG) for project 448588364 of the Emmy Noether Programme.
The authors thank the International Max Planck Research School for Intelligent Systems (IMPRS-IS) for supporting Tim Z.~Xiao.

\paragraph{Reproducibility Statement.}
All code necessary to reproduce the results in this paper is available at \href{https://github.com/timxzz/ABI_with_Rebasin}{https://github.com/timxzz/ABI\_with\_Rebasin}.

\bibliographystyle{plain}
\bibliography{ref}

\newpage
\appendix

\section{Experiment Setup}

We use a fully connected network with a single hidden layer of size 512 for all models in MNIST classification setting.
More specifically, for deep ensemble, we use 5 randomly initialized members, trained with Adam for 50 epochs using a maximum a posteriori (MAP) with standard Gaussian prior.
For VI, the network has double the size of parameters since we need to model both the mean and variance.
It is also trained with Adam for 50 epochs.
When evaluating their predictive distribution, for both ensemble and VI, we use 100 samples in Monte Carlo integration.

For HMC, we use 600 epochs for burn-in, then we save a sample for every 10 epochs.
Each epoch uses 500 leapfrog steps.
In total, we generated 1000 HMC samples for the evaluations in the paper.

\end{document}